\documentclass[letterpaper, 10 pt, journal, twoside]{ieeetran}

\usepackage{graphicx} 
\usepackage{amsmath}
\usepackage{amssymb}

\usepackage{amsthm}
\usepackage{svg}
\usepackage{caption}
\usepackage{subcaption}
\usepackage{xcolor}
\usepackage{array}
\usepackage{flushend}
\usepackage{multirow, makecell}
\usepackage{dblfloatfix}
\usepackage{tabularx}
\usepackage{booktabs}
\usepackage{algorithm}
\usepackage{algpseudocode}
\usepackage{setspace}
\usepackage{dcolumn}
\usepackage{bm}       
\usepackage{lipsum} 
\usepackage{cite}
\usepackage{url}
\usepackage{pdflscape}
\usepackage{rotating}

\IEEEoverridecommandlockouts

\makeatletter
\newcommand*{\@rowstyle}{}

\newcommand*{\rowstyle}[1]{
  \gdef\@rowstyle{#1}%
  \@rowstyle\ignorespaces%
}

\newcolumntype{=}{
  >{\gdef\@rowstyle{}}%
}

\newcolumntype{+}{
  >{\@rowstyle}%
}
\makeatother
\newcommand{\RomanNumeralCaps}[1]
    {\MakeUppercase{\romannumeral #1}}

\newcolumntype{d}[1]{D{.}{.}{#1}}

\newcommand{\revisedtext}[1]{#1}
\newcommand{\reviewedtext}[1]{#1}
\newcommand{\reviewed}[1]{#1}
\newcommand{\revised}[1]{#1}

\usepackage[bookmarks=true,colorlinks]{hyperref}
\hypersetup{citecolor=blue}

\DeclareMathOperator*{\argmax}{argmax}

\title{The Harmonic Exponential Filter for Nonparametric Estimation on Motion Groups}
\author{Miguel Saavedra-Ruiz, Steven A. Parkison, Ria Arora, James Richard Forbes, and Liam Paull
\thanks{Miguel Saavedra-Ruiz, Steven A. Parkison, Ria Arora, and Liam Paull are with the Department of Computer Science and Operations Research, Université de Montréal, Montréal, QC, Canada, and with Mila - Quebec AI Institute, Montréal, QC, Canada.}
\thanks{James Richard Forbes is with the department of Mechanical Engineering, McGill University, Montréal, QC, Canada.}
\thanks{This research was partially funded by an FRQNT B2X Scholarship [MS]. We acknowledge the support of the Natural Sciences and Engineering Research Council of Canada (NSERC), [funding reference numbers RGPIN-2024-04956 and RGPIN-2023-03621], as well as CIFAR under the CCAI Chairs program.} 
}

\begin{document}

\begin{minipage}{\textwidth}

\begin{center}
\vspace*{\fill}
\textcopyright 2025 IEEE.  Personal use of this material is permitted.  Permission from IEEE must be obtained for all other uses, in any current or future media, including reprinting/republishing this material for advertising or promotional purposes, creating new collective works, for resale or redistribution to servers or lists, or reuse of any copyrighted component of this work in other works. 
\vspace*{\fill}
\end{center}

\hfill

This paper has been accepted for publication in the \textit{IEEE Robotics and Automation Letters}. \\ Please cite this paper as:

\begin{verbatim} 
@article{saavedra2025hef,
    title        = {The Harmonic Exponential Filter for Nonparametric Estimation on 
                    Motion Groups},
    author       = {Saavedra-Ruiz, Miguel and Parkison, Steven A. and Arora, Ria and 
                    Forbes, James Richard and Paull, Liam},
    year         = 2025,
    journal      = {IEEE Robotics and Automation Letters},
    volume       = {},
    number       = {},
    pages        = {1--8},
    doi          = {10.1109/LRA.2025.3527346}
}   \end{verbatim} 
\end{minipage}
\newpage

\maketitle

\begin{abstract}
Bayesian estimation is a vital tool in robotics as it allows systems to update the robot state belief using incomplete information from noisy sensors. To render the state estimation problem tractable, many systems assume that the motion and measurement noise, as well as the state distribution, are unimodal and Gaussian. However, there are numerous scenarios and systems that do not comply with these assumptions. Existing nonparametric filters that are used to model multimodal distributions have drawbacks that limit their ability to represent a diverse set of distributions. This paper introduces a novel approach to nonparametric Bayesian filtering on motion groups, designed to handle multimodal distributions using harmonic exponential distributions. This approach leverages two key insights of harmonic exponential distributions: a) the product of two distributions can be expressed as the element-wise addition of their log-likelihood Fourier coefficients, and b) the convolution of two distributions can be efficiently computed as the tensor product of their Fourier coefficients. These observations enable the development of an efficient and asymptotically exact solution to the Bayes filter up to the band limit of a Fourier transform. We demonstrate our filter's performance compared with established nonparametric filtering methods across simulated and real-world localization tasks. \footnote{\label{note1}Project page \url{https://montrealrobotics.ca/hef/}}

\end{abstract} 
\section{Introduction}

Recursive Bayesian filtering is a powerful approach for state estimation in robotics. It allows the fusion of different sensor streams while producing \revisedtext{belief estimations over the state} that can be used in downstream tasks such as planning, control, and mapping. To render the state estimation problem computationally tractable, certain assumptions are frequently invoked, including the presumption that the state distribution, as well as the motion and measurement noise, are unimodal and Gaussian distributed. \revisedtext{However, there are many scenarios where a unimodal Gaussian belief fails to place a high probability (density) over the true state}. Estimates involving uncertain data associations \cite{thrun2005probrob}, propagating errors through rigid-body motions \cite{long2012banana}, and measurements that suffer from multipath errors \cite{dardari2009multipath} and symmetries \cite{xiang2018posecnn} all result in distributions that are distorted or multimodal.

Gaussian filters have been extended to handle some non-Gaussian distributions by modeling the error of the state in Lie groups using parametric distributions \cite{wang2008nonparametric, barreau2017iekf}. However, to propagate errors, these methods rely on approximations or parametric assumptions to maintain a tractable solution. 

\begin{figure}[t]
\centering
\begin{subfigure}[b]{0.49\columnwidth}
\includegraphics[width=\textwidth]{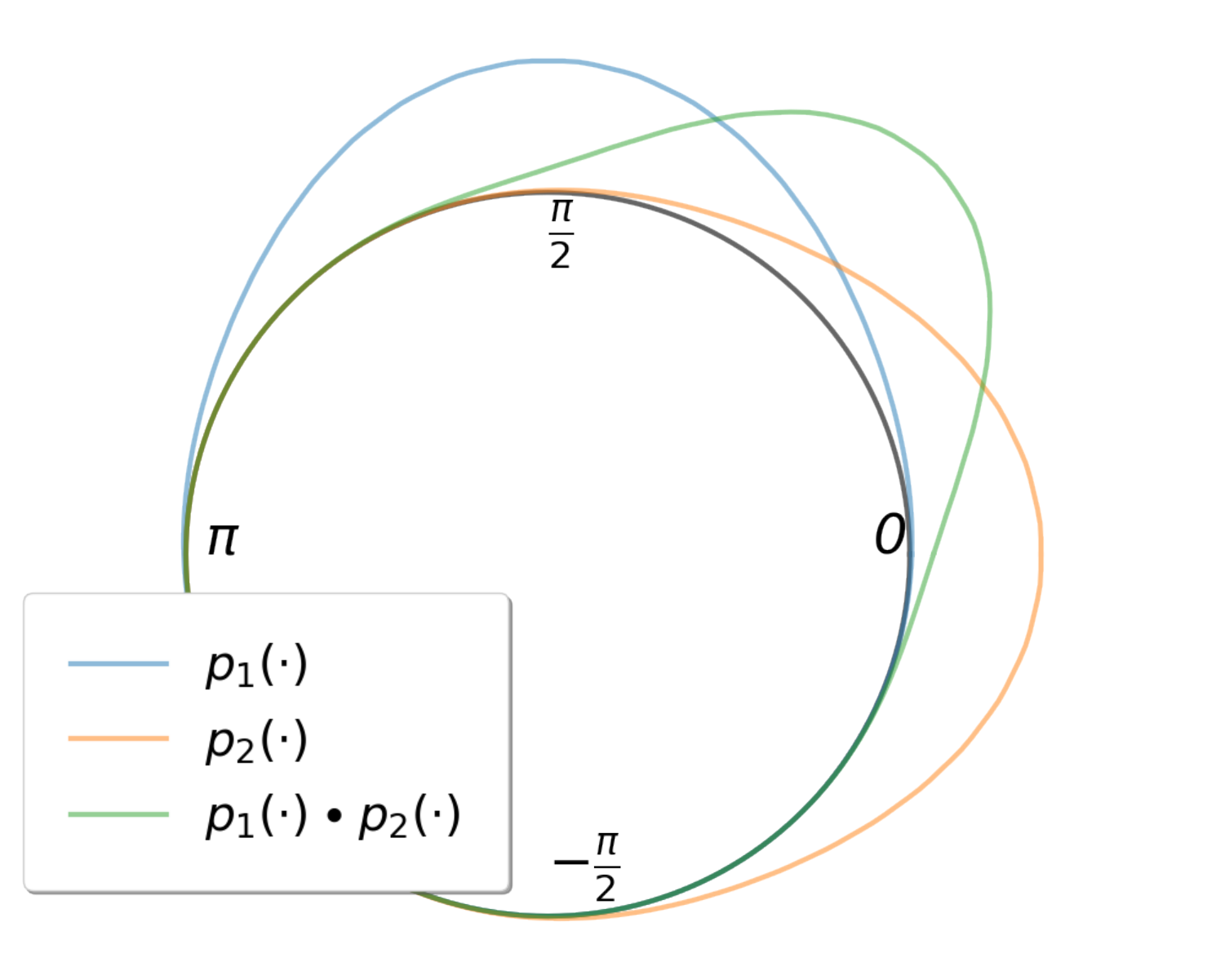}
\caption{Product}
\end{subfigure}
\begin{subfigure}[b]{0.49\columnwidth}
\includegraphics[width=\textwidth]{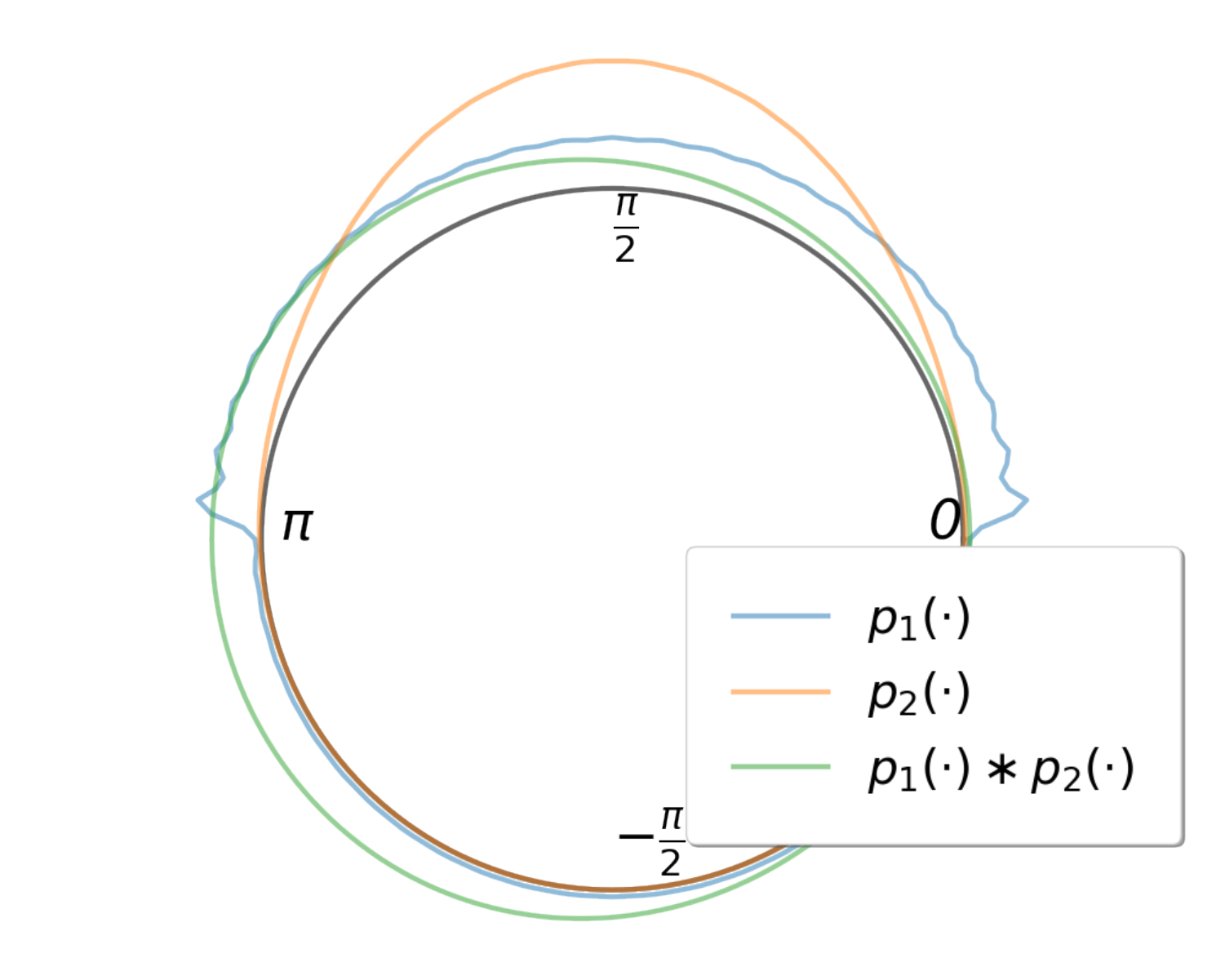}
\caption{Convolution}
\end{subfigure}
\caption{The product (a) and convolution (b) operations between two distributions play an important role in recursive Bayesian filtering. The examples here are shown with harmonic exponential distributions on the circle $S^1$. }
\label{fig:prod_conv}
\end{figure}

Nonparametric filtering offers an alternative to Gaussian-based filters, such as particle filters \cite{pf, li2013adapting, elvira2017improving} and histogram filters \cite{histogram, zhou2021quaternion, li2021grid}. However, these approaches are not free of limitations. Particle filters are susceptible to particle deprivation \revisedtext{\cite[Chapter 4]{thrun2005probrob}}, where the correct state is no longer represented within the particle set. Similarly, histogram filters require resource-intensive computations, including convolutions during motion propagation, and are only accurate up to the quantization error when computing probability bins. 

\reviewed{This paper draws on seminal ideas in Fourier analysis for optimal filtering on rotation groups \cite{willsky1974circle1, willsky1974circle2, lo1979exponential, marcus1979harmonics} and proposes a novel nonparametric, multimodal filtering approach for motion groups, leveraging harmonic exponential distributions \cite{cohen2015harmonic}.} We name this method the harmonic exponential filter (HEF) and it relies on three key benefits of harmonic exponential distributions. First, the prediction step exploits the Fourier convolution theorem, yielding an efficient tensor product instead of a convolution during motion propagation. Second, the update step of the filter becomes an element-wise addition. Finally, the Fourier coefficient representation of the harmonic exponential distribution represents more naturally smoothly-varying \revisedtext{density functions} when compared to histogram binning. These steps are illustrated in Fig~\ref{fig:prod_conv}. 

In summary, our main contributions are:

\begin{itemize}
    \item An \reviewed{asymptotically} exact Bayesian filter that operates on a compact subset of motion groups, for band-limited prior, motion, and measurement functions.
    \item A nonparametric filtering approach for noncommutative motion groups that captures multimodal distributions.
    \item An efficient motion step implementation for motion groups that leverages the Fourier convolution theorem.
    \item \reviewed{An experimental validation of our approach against classical filtering methods.}
\end{itemize}

We show that our method outperforms the baselines in three non-Gaussian filtering tasks using one simulated and two real-world datasets.

\section{Related Work} \label{sec:rw}

Various approaches have extended Bayesian filtering beyond the unimodal Gaussian assumptions of the Kalman filter (KF). For instance, the KF has been adapted to operate with a mixture of parameterized distributions instead of a unimodal Gaussian \cite{sorenson1971sum}. Gilitschenski \emph{et al.} \cite{gilitschenski2019deep} adapted a mixture model to use direction statistics on the manifold of rotation groups. However, a key challenge with such approaches is the growth of mixture components with each iteration. Common approaches to overcome this are based on distance metrics to merge mixture components \cite{schoenberg2009sumfilters} or resampling \cite{chen2000mixture}.

The KF has also been extended to exploit \emph{symmetries} of the system by characterizing the state of the robot using Lie groups \cite{sola2018micro}. If the state error is modeled in the Lie algebra of motion groups with a Gaussian \revisedtext{belief} and then mapped back to the group with the exponential map, a non-Gaussian \revisedtext{posterior} is obtained on the manifold \cite{wang2008nonparametric, long2012banana}. However, to propagate the error state over time and attain a tractable solution, these methods rely on second-order approximations. \reviewed{The invariant extended Kalman filter (IEKF) \cite{barreau2017iekf} also models the state error in the Lie algebra with a Gaussian density, which restricts it to unimodal distributions.} 

Unlike the various KF approaches, our method is nonparametric and can fit any distribution in a broader family of distributions while preserving system symmetries. A well-known nonparametric filtering approach is the histogram filter, which approximates the posterior \revisedtext{belief} by discretizing the state space and assigning a probability value to each bin \cite{histogram}. These methods can be adapted to the topology and action of the group \cite{kurz2018hist} and have been widely studied to model distributions on rotation groups \cite{zhou2021quaternion, li2021grid}. Our method resembles histogram filters as we require discrete samples of the manifold to represent the posterior. However, our formulation operates in both rotation and motion groups and is more computationally efficient during the propagation step than histogram filters. What is more, our approach is exact up to the band limit of a discrete Fourier transform. 

Another family of nonparametric and multimodal filters is the particle filter which represents the posterior belief using a discrete set of particles \cite{pf}. These methods can model general distributions, but are known to suffer from particle deprivation, particularly in high-dimensional state spaces \revisedtext{\cite[Chapter 4]{thrun2005probrob}}. Research efforts have been devoted to addressing this problem by proposing better sampling schemes \cite{li2013adapting, elvira2017improving}, \reviewedtext{and by changing how particles evolve in the prediction step to better match the posterior distribution\cite{steinpf}}. In contrast, our method does not use particles to model the posterior \revisedtext{belief} and does not suffer from particle deprivation.

Kernel density estimates are a prevalent method to represent nonparametric \revisedtext{probability density functions}. They can be adapted to handle belief propagation through probabilistic graphical models~\cite{sudderth2010nonparametric}, and the product of distributions within a filtering framework \cite{song2013kde}. \reviewed{Similarly, learning-based methods have been used to approximate sampling from a non-Gaussian posterior in SLAM by modeling conditional distributions in a Bayes tree with normalizing flows \cite{huang2022incremental}.} 

Harmonic analysis for optimal filtering has been widely studied on rotation groups. Willsky \cite{willsky1974circle1} derived state estimation equations for filtering on the circle using infinite Fourier coefficients and proposed finite-dimensional suboptimal implementations \cite{willsky1974circle2}. Filtering with ``Fourier densities'' has since expanded to groups like $SO(N)$ \cite{marcus1979harmonics}. Similarly to our work, Lo~\emph{et al.} \cite{lo1979exponential} highlighted the advantages of rotational Fourier densities for the update step on $SO(3)$. Fourier densities have also been applied to non-linear filtering, with Brunn~\emph{et al.} \cite{brunn2006efficient, brunn2006nonlinear} formulating Bayesian prediction and update steps on d-dimensional Euclidean spaces. Unlike these methods, our approach extends these ideas beyond compact rotation groups or Euclidean spaces and applies them to motion groups.

Pfaff~\emph{et al.} \cite{plaff2015fourierfilt} proposed a method conceptually similar to ours, called Fourier filtering, which was applied to the unit circle and sphere \cite{plaff2017fourierfilt, pfaff2019fourier}. \reviewed{They also showed their method applied to the problem of multimodal density estimation on the hypertorus, and addressed arbitrary transition densities in the prediction step \cite{pfaff2016multivariate, pfaff2016nonlinear}.} 

Although \reviewed{these approaches} resemble ours in that probability \revisedtext{densities} are represented by Fourier coefficients, our approach extends this concept in several meaningful ways. First, since we model \revisedtext{density functions} in the log-density space, the product of distributions becomes an element-wise addition instead of the more \revisedtext{intricate} tensor product needed for Fourier filtering. Second, by representing distributions with the Fourier coefficients derived from the log-density function, the corresponding probability \revisedtext{density function} is guaranteed to be non-negative \reviewed{\cite{duran2004circular}}. Lastly, our approach can represent densities on \reviewed{compact subsets} of motion groups $SE(N)$, particularly any matrix Lie group and not just on a circle or a sphere \reviewed{(the latter of which is not a \revised{Lie} group)}.

\section{Harmonic Exponential Filtering} \label{sec:method}

Fourier analysis provides a way to parameterize \revisedtext{well-behaved functions in $\mathbb{R}^N$ (i.e. one that is square integrable $\mathcal{L}^2$ on $\mathcal{V} \subseteq \mathbb{R}^N$ and has a finite number of discontinuities in any finite interval in $\mathbb{R}^N$)} using an infinite sum of trigonometric basis functions. This approach has proven to be useful for various engineering applications for several reasons. Moreover, the convolution theorem shows that computing integrals in spatial domains can be replaced by an efficient tensor product in the frequency domain. A generalized form of Fourier analysis can be used on \reviewed{compact subsets of} motion groups. The topic is presented in depth by Chirikjian and Kyatkin in \revisedtext{\cite[Chapters 8,10]{chirikjian2001engineering}}.
This section presents an overview of our approach. First by \reviewedtext{providing a brief introduction to} Fourier analysis on motion groups, \reviewed{particularly $SE(2)$} (Sec.~\ref{ssec:hamg}), then showing how it is used in the family of harmonic exponential distributions to represent a varied set of densities (Sec.~\ref{ssec:hed}), and finally how these concepts can be combined to perform Bayesian filtering (Sec.~\ref{ssec:bayesfilter}).

\subsection{Harmonic Analysis on Motion Groups} \label{ssec:hamg}

Motion groups are a specific category of Lie groups \cite{sola2018micro} that represent rigid-body motions within the Euclidean space $\mathbb{R}^N$. This classification includes other important subgroups, such as rotation groups $SO(N)$ and \reviewed{rigid-body-transformation} groups $SE(N)$. Particularly, these are \revisedtext{Lie} groups because they \revisedtext{satisfy the group axioms (see \cite[Chapter 7]{chirikjian2001engineering})} \reviewedtext{and are a differentiable manifold}. A group consists of a set \( G \) and a binary operator \( \circ \) that satisfies the closure property; \(\forall g, h \in G \), their composition \( g \circ h \) also belongs to \( G \).

Conducting Fourier analysis on \reviewedtext{functions supported on} \reviewed{Lie groups, such as $SE(2)$}, requires a collection of carefully chosen basis functions. \reviewedtext{These basis functions are in turn the matrix elements of} \emph{irreducible unitary representations} (IURs). \reviewedtext{A definition of IURs is provided below.}

\reviewed{A matrix \emph{representation} for $SE(2)$, denoted $R^\lambda(g)$, is an infinite-dimensional matrix function that satisfies the homomorphism property $R^\lambda(g\circ h)=R^\lambda(g)R^\lambda(h)$, $\forall g,h \in G = SE(2)$, where \( \lambda \in \mathbb{R}^+ \) indexes the representation. Here, $g\circ h$ represents composition of group elements, while $R^\lambda(g)R^\lambda(h)$ denotes standard matrix multiplication.}

\reviewed{A \emph{unitary representation} \( U^\lambda_{mn}(g) \), with matrix elements $m, n \in \mathbb{Z}$, is a representation that satisfies $\langle U^\lambda(g)\varphi_1, U^\lambda(g)\varphi_2  \rangle = \langle \varphi_1, \varphi_2 \rangle$ for all $m, n \in \mathbb{Z}$. Here, the inner product is defined as \( \langle \varphi_1, \varphi_2 \rangle = \frac{1}{2\pi} \int_0^{2\pi} \varphi_1(\psi)^{\mathsf{H}} \varphi_2(\psi) d_\psi\) where $(\cdot)^{\mathsf{H}}$ denotes the Hermitian transpose and $\varphi_i$ is a function defined on $S^1$. An \emph{irreducible unitary representation} (IUR) is a unitary representation that cannot be block-diagonalized further into another representation. A representation can be block-diagonalized if it is equivalent to the direct sum of other representations \cite{chirikjian2001engineering}. An IUR is represented as $T^\lambda(g) \in \mathbb{R}^{m \times n}$.}

IURs are the foundation for performing harmonic analysis on \reviewed{motion} groups. If we have a function, $f: G \rightarrow \mathbb{R}$, that is square integrable, and an IUR of \reviewed{motion group} $G$, $T^\lambda(G)$, then we can perform a generalized Fourier transform to compute the frequency components of $f$ as

\begin{equation}
\begin{aligned}\label{eq:fft}
\mathcal{F}[f]_{mn}^\lambda = \int_G f(g) T_{mn}^\lambda(g)^{\mathsf{H}} d(g) = \eta_{mn}^\lambda , \\   
\end{aligned}
\end{equation}
where $d(g)$ is the bi-invariant volume
element of $G$. Matrix elements of the Fourier coefficients \reviewedtext{$\eta^\lambda \in \mathbb{R}^{m \times n}$, and IUR $T^\lambda(g) \in \mathbb{R}^{m \times n}$ are indexed by their row and column indices} $m$ and $n$. \reviewed{Exploiting the fact that $T(g)$ is a unitary representation, the generalized inverse Fourier transform is} 
\begin{equation}
\begin{aligned}
\label{eq:invfft}
\mathcal{F}^{-1}[\eta](g) =  \sum_{n,m\in\mathbb{Z}} \int_0^\infty \eta_{mn}^\lambda T_{nm}^\lambda(g) \lambda d_\lambda = f(g), \\
\end{aligned}
\end{equation}
\noindent
\reviewed{with $\mathcal{F}\left[\mathcal{F}^{-1}[\eta]\right] = \eta$ and $\mathcal{F}^{-1}\left[\mathcal{F}[f]\right] = f$. Chirikjian and Kyatkin provide a proof of the Fourier inversion formula for $SE(2)$ in \cite[Section 10.4.2]{chirikjian2001engineering} and offer a detailed explanation of the fast Fourier transform (FFT) over 2D motion groups in \cite[Section 11.2]{chirikjian2001engineering}.}

In this work we use the IUR of $SE(2)$ presented in \cite{chirikjian2001engineering,chirikjian2003se2fft}. For \reviewedtext{$g(x, y, \theta) \in SE(2)$} with $(x,y) \in \mathbb{R}^2$ and $\theta \in [0, 2\pi]$ parameterizing an element of $SE(2)$, \reviewedtext{a variable $\psi \in [0, 2 \pi)$, bi-invariant volume $d\mu \triangleq dx \, dy \, d\theta \, d\psi$ and complex unit $i$}, the \reviewedtext{Fourier coefficients} $\eta^\lambda_{mn}$ are computed as 
\begin{equation*}
\begin{split}
\eta_{mn}^\lambda = 
\int_{x,y,\theta,\psi} f(x, y, \theta)
e^{i\psi n}
e^{i(x \lambda \textrm{cos}\psi+y \lambda \textrm{sin}\psi)}
e^{-im(\psi-\theta)} d\mu.
\end{split}
\end{equation*}

\reviewed{To make the integral in \eqref{eq:fft} tractable, we assume that $f$ is supported on compact subsets of the group and use a discrete subset of function samples. For \( SE(2) \), the orientation component \( SO(2) \) is compact, and for translation, we consider a predefined compact subset of \( \mathbb{R}^2 \), as described in \cite{chirikjian2001engineering}. Additionally, we assume $f$ is bandlimited with $\lambda \in (0, \lambda_{\text{max}}]$ and truncate the infinite sums in \eqref{eq:invfft} to $-k \leq m, n \leq k$. We will omit indices and denote the generalized Fourier transform for \reviewed{compact subsets of} motion groups as $\eta = \mathcal{F}[f]$.}

\subsection{Harmonic Exponential Distributions}\label{ssec:hed}

\begin{figure}[t]
\vspace{0.65em}
\centering
\includegraphics[width=0.7\columnwidth]{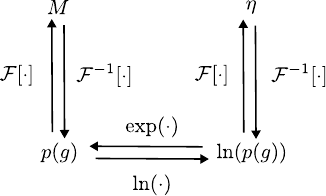}
\caption{The relationship between probability density $p(g)$, log-density $\text{ln}(p(g))$, and their spectral coefficients $\eta$ and $M$. \revisedtext{We model the log-density $\text{ln}(p(g))$ using $\eta$. To recover the $p(g)$, we compute the inverse FFT of $\eta$ and exponentiate the result. The spectral coefficients $M$ are obtained by taking the FFT of $p(g)$ and are used to compute convolutions.}}
\vspace{-1.25em}
\label{fig:spaces}
\end{figure}

Probability distributions, which must abide by the probability axioms, are non-negative functions, $p(g) \ge 0, \forall g \in G$, \revisedtext{which} integrate to 1, $\int_G p(g) dg = 1$. Cohen and Welling~\cite{cohen2015harmonic} presented the harmonic exponential distribution \revisedtext{as a class of probability distributions supported on groups and homogeneous spaces}. \revisedtext{Its density function is defined as}
\begin{equation}\label{eq:hed}
p_\eta(g) \triangleq p(g \mid \eta) = \frac{1}{Z_\eta} \text{exp}(\eta \cdot T(g)),    
\end{equation}
\noindent
where $\eta$ are the \reviewed{(spectral)} parameters, $T(g)$ is the IUR of group $G$, $Z_\eta$ is the normalizing constant, and $(\cdot)$ is \reviewed{a shorthand notation for $\eta \cdot T(g) \triangleq \mathcal{F}^{-1}[\eta](g)$. Both notations will be used interchangeably to remain consistent with \cite{cohen2015harmonic}.} 
The parameters $\eta$ and $Z_\eta$ can be computed via harmonic analysis from $\phi(g \mid \mathbf{\eta}) \triangleq \exp(\mathbf{\eta} \cdot T(g))$ as
\begin{subequations}
\label{eq:ll}
\begin{align}
\ln\phi(g \mid \eta) &= \eta \cdot T(g) = \mathcal{F}^{-1}[\eta](g), \label{eq:lla} \\
Z_\eta &= \int_G \phi(g \mid \eta) T^{0}_{00}(g) \, d(g) =  \mathcal{F}[\phi]^{0}_{00}. \label{eq:llb}
\end{align}
\end{subequations}
\noindent
with $\mathcal{F}[\phi]^{0}_{00}$ being the zeroth Fourier coefficient and $T^{0}_{00}(g) = 1$. Representing the probability density with Fourier coefficients in log-density space, ensures that the probability \revisedtext{density} function is non-negative \revisedtext{for any set of parameters $\eta$}, and the normalizing constant $Z_\eta$ ensures that it integrates to 1. \revisedtext{From this point forward, we use the shorthand notation $p_a(g) \triangleq p_{\eta_a}(g)$.} The harmonic exponential distribution provides an elegant solution for the product of two distributions, achieved through the element-wise addition of their Fourier coefficients:
\begin{equation}\label{eq:prod}
\begin{aligned}
p_a (g) p_b (g) & = \frac{1}{Z_\eta^a Z_\eta^b}\exp((\eta_a + \eta_b) \cdot T(g))  \\
& = p(g \vert \eta_a + \eta_b).
\end{aligned}
\end{equation}

Convolving two distributions is inherently more \revisedtext{complicated}. However, harmonic analysis provides a connection between convolutions and spectral coefficients,
\begin{equation}\label{eq:conv}
\begin{aligned}
(p_a \ast p_b) (g) &= \int_G p_a(h) p_b(h^{-1} \circ g) d(h) \\
& = \mathcal{F}^{-1}[\mathcal{F}[p_b] \mathcal{F}[p_a]](g) = \mathcal{F}^{-1}[M_b M_a](g) \\
\end{aligned}
\end{equation}
with $ \mathcal{F}[p_a] = \int_{G} p_a(g) T(g) d(g) = M_a$ being the Fourier coefficients of the density $p_a$, instead of a log-density function as in \eqref{eq:lla}. Due to the nonconmutative nature of \reviewed{$SE(2)$}, encoded through $T(g)$, the order of the Fourier coefficients is reversed in~\eqref{eq:conv}. Note that $M_a$ and $M_b$ are composed via matrix products: $M^\lambda = M_b^\lambda M_a^\lambda$ \cite[Section 10.4]{chirikjian2001engineering}. The relation between these spaces is shown in Fig.~\ref{fig:spaces}.

\subsection{Bayesian Filtering}\label{ssec:bayesfilter}

Following the notation of Thrun \emph{et al.}~\cite[Chapter 2]{thrun2005probrob},
equations~\eqref{eq:prod} and \eqref{eq:conv} allow us to incorporate the harmonic exponential distribution into a Bayesian filter:
\begin{equation}\label{eq:bayes}
\begin{aligned}
bel(x_t) \triangleq p(x_t\vert z_{1:t},u_{1:t}) &= \frac{p(z_t \vert x_t)p(x_t \vert z_{1:t-1}, u_{1:t})}{p(z_t \vert z_{1:t-1}, u_{1:t})} \\
&= \gamma p(z_t \vert x_t)\overline{bel}(x_{t}),
\end{aligned}
\end{equation}

\noindent
\revisedtext{where the predicted belief over the state $x_t$ after incorporating the motion model but before incorporating the measurement $z_t$ is denoted as $\overline{bel}(x_{t}) \triangleq p(x_t \vert z_{1:t-1}, u_{1:t})$, and the posterior belief after incorporating $z_{t}$ as $bel(x_t)$.}

The prediction step involves defining the motion model, which represents the likelihood of the current pose \( x_t \in SE(N) \) given the previous pose \( x_{t-1} \in SE(N) \) and control input \( u_t \), i.e., \( p(x_t \vert u_t, x_{t-1}) \). \reviewed{Defining the motion model as $x_t = x_{t-1} \circ u_t$, where $u_t$ is a random variable with a known density, allows us to compute the relative motion as $x^{-1}_{t-1} \circ x_t = u_t$. Therefore, the relative motion likelihood is $p_{u_t} (x^{-1}_{t-1} \circ x_t)$, giving us the prediction step}
\begin{equation}\label{eq:predict}
\begin{aligned}
\overline{bel}(x_t) &= \int_{SE(N)} p(x_t \vert u_t, x_{t-1}) bel(x_{t-1}) \, dx_{t-1} \\
&= \int_{SE(N)} bel(x_{t-1}) p_{u_t} (x^{-1}_{t-1} \circ x_t) \, dx_{t-1} \\ 
&= \mathcal{F}^{-1}[\mathcal{F}[p_{u_t}] \mathcal{F}[bel_{t-1}]](x_{t}).
\end{aligned}
\end{equation}

The update step of a Bayes' filter is the product of two probability distributions, $p(z_t \vert x_t) \overline{bel}(x_t)$. \revisedtext{If we perform Fourier analysis in the domain $x_t$ of the measurement log-likelihood function we can get a set of Fourier coefficients $\eta_{z_t \vert x_t} \triangleq \mathcal{F}[\text{ln}(p(z_t\vert x_t)]$, and the measurement likelihood function becomes $p(z_t|x_t) = 1 / Z_{\eta}\exp(\eta_{z_t\vert x_t} \cdot T(x_t))$. With that and using \eqref{eq:prod}, the update step simplifies to}
\begin{equation}\label{eq:udpate}
bel(x_t) =  \gamma \exp \left((\eta_{z_t | x_t} + \eta_{b\overline{el}_t}) \cdot T(x_t)\right),
\end{equation}
where the normalizing constant $\gamma$ is computed from the zeroth Fourier coefficient as shown in~\eqref{eq:llb}. 

We use $bel_t: SE(N) \to \mathbb{R}^{\geq  0}$ to represent the probability density function for the posterior at time $t$. The predicted belief $\overline{bel}_t$, motion model \( p_{u_{t}} \), and measurement likelihood $p_{z_{t} \vert x_{t}}$ are similarly denoted. Note that these definitions are functions, not point estimates, allowing us to compute Fourier coefficients for our density functions as $\eta_{z_t \vert x_t} \triangleq \mathcal{F}[\ln (p_{z_t \vert x_t})]$ and $\eta_{\overline{bel}_t} \triangleq \mathcal{F}[\ln (\overline{bel_t})]$. The harmonic exponential Bayes filter is presented in Algorithm~\ref{algo:algorithm}.

\begin{algorithm}[!hptb]
\caption{The Harmonic Exponential Bayes Filter}
\label{algo:algorithm}

\hspace*{\algorithmicindent} \textbf{Input}: Previous posterior belief $bel_{t-1}$, motion likelihood $p_{u_t}$, measurement likelihood $p_{z_{t} \vert x_{t}}$ and IUR $T$.
\begin{algorithmic}[1]
\setstretch{1.2}
\Statex \textcolor{gray}{// Prediction Step}
\State $\overline{bel}(x_t) = \mathcal{F}^{-1}[\mathcal{F}[p_{u_t}] \mathcal{F}[bel_{t-1}]](x_{t}) $ 
\Comment{Eq.~\eqref{eq:predict}}
\Statex \textcolor{gray}{// Compute Fourier Coefficients}
\State $\eta_{z_t \vert x_t} \gets  \mathcal{F}[\ln (p_{z_t \vert x_t})]$
\Comment{Eq.~\eqref{eq:lla}}
\State $\eta_{b\overline{el}_t} \gets  \mathcal{F}[\ln (\overline{bel_t})]$
\Comment{Eq.\eqref{eq:lla}}
\Statex \textcolor{gray}{// Update Step}
\State \reviewed{$\gamma^{-1} \gets \mathcal{F}\left[ \exp \left((\eta_{z_t \vert x_t} + \eta_{b\overline{el}_t}) \cdot T(x_t)\right) \right]_{00}^0$}
\Comment{Eq.~\eqref{eq:llb}}
\State $bel(x_t) = \gamma \exp \left((\eta_{z_t \vert x_t} + \eta_{b\overline{el}_t}) \cdot T(x_t)\right)$ 
\Comment{Eq.~\eqref{eq:udpate}}
\State \Return $bel_t$
\end{algorithmic}
\end{algorithm}
\vspace{-0.75em}

\subsection{Analysis of the Harmonic Exponential Filter} \label{ssec:expressivity}

We now point out several advantages inherent to the HEF. One is efficiency compared to a histogram-based approach. The HEF requires a transition between two different Fourier spaces to compute a convolution, which inherently adds complexity. Nevertheless, it can still be more efficient than a standard convolution implementation in 3D.
\revised{If we consider a discretized space of $n_s\times n_s\times n_s$ in 3D, a convolution has complexity $O(n_s^6)$, while a $SE(2)$ Fourier transform using FFT methods is of $O(n_s^3\log{n_s^3})$. The previous calculation assumes $m, n \approx n_L \approx n_s$ and \( \lambda \approx n_\lambda \approx n_s \). Under these assumptions, the tensor product needed for \eqref{eq:conv} is $O(n_\lambda n_L^3)$ and the complexity of \eqref{eq:invfft} is also $O(n_s^3\log{n_s^3})$.}

\revisedtext{Table~\ref{table:runtime4} shows a runtime comparison between a histogram-based approach and the HEF over varying sample grid sizes. As shown, the histogram approach quickly becomes intractable. A workaround we use for the benchmark algorithm in Sec.~\ref{sec:eval} is to convolve with a smaller kernel, eliminating areas with near zero probability, but that is only an option for certain densities, such as low covariance Gaussians.}

\begin{table}[!htpt]
\vspace{0.65em}
\medskip
\centering
\begin{small}
\begin{tabular}{=l+r+r+r}
\toprule
& \multicolumn{3}{c}{Convolution Runtime (s)} \\
\cmidrule{2-4} 
 & \multirowcell{2}{Grid size \\ $(10, 10, 8)$} & \multirowcell{2}{Grid size \\ $(20, 20, 8)$} & \multirowcell{2}{Grid size \\ $(40, 40, 8)$} \\ \\
\midrule
Hist (PP) & 0.2284 & 15.7969 & 997.2984  \\
Hist (Scipy) & 0.0375 & 2.2434 & 146.4707 \\
HEF (Ours) & \textbf{0.0014} & \textbf{0.0083} & \textbf{0.0684} \\
\bottomrule
\end{tabular}
\end{small}
\caption{Runtime comparison of a histogram-based approach and the HEF for convolving two functions. Hist (PP) is a direct convolution written in pure python, Hist (Scipy) uses \texttt{scipy.signal.convolve()}, and HEF (ours) implements \eqref{eq:conv} in Python using the NumPy FFT library.}
\label{table:runtime4}
\vspace{-1.25em}
\end{table}

\begin{figure}[!htpb]
\vspace{-1.0em}
\centering
\begin{subfigure}[b]{0.6\columnwidth}
\includegraphics[width=\textwidth]{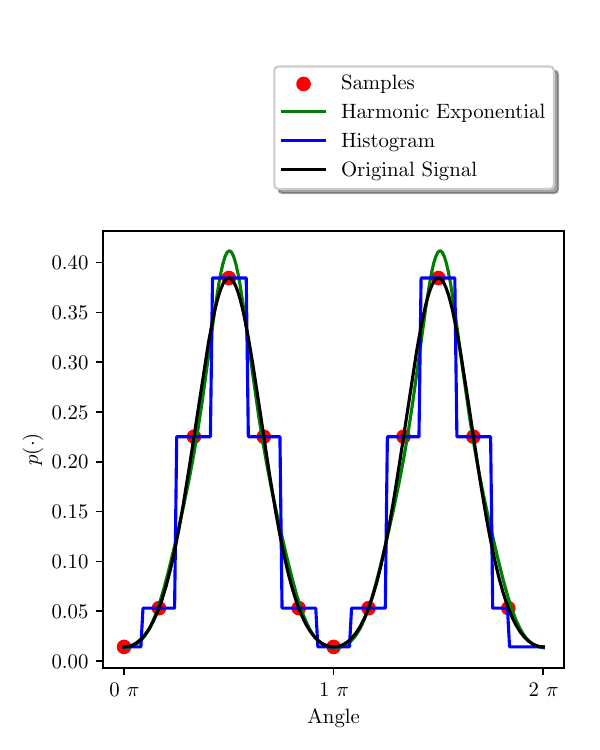}
\vspace{-2em}
\subcaption{}
\end{subfigure} 
\hspace{1em}
\begin{subfigure}[b]{0.3\columnwidth}
\includegraphics[width=\textwidth]{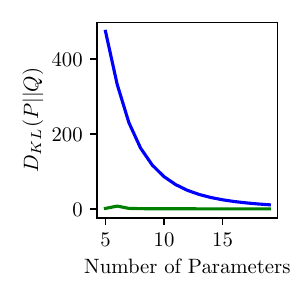}
\includegraphics[width=\textwidth]{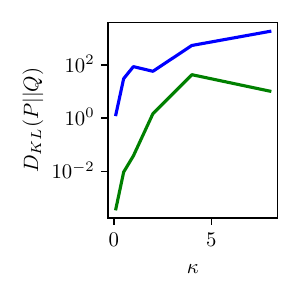}
\caption{}
\end{subfigure}
\caption{Analysis of the harmonic exponential distribution and histogram distributions in modeling a mixture of two von Mises distributions on $S^1$. (a) depicts how each method reproduces the underlying mixture.  (b) shows how $D_{KL}$ evolves for each method as the number of parameters used increases, and shows the effect of the concentration parameter $\kappa$ on the fidelity of each method, as measured by $D_{KL}$.
}
\label{fig:von_mise}
\vspace{-0.5em}
\end{figure}

Another advantage of representing \revisedtext{density functions} with Fourier coefficients over probability bins in histogram filters is the fidelity with which they can represent signals. When computing the Fourier coefficients for the harmonic exponential distribution, FFT operations behave as a low-pass filter on the signals, leading to information loss in the distribution. However, binning a probability distribution also causes a loss of information. To compare the two, we computed the Kullback–Leibler (KL) divergence $D_{KL}(P \vert \vert Q) = \int_{\mathcal{X}} p(x) \log \left(\frac{p(x)}{q(x)}\right)dx$, between the two approaches and a mixture of two von Mises distributions on \reviewed{the Lie group $S^1$}. This analysis is shown in Fig.~\ref{fig:von_mise}, where the KL divergence is used as a measure of distance between two distributions.

The von Mises distribution is a probability distribution on the circle $S^1$ whose density function is given by 

\begin{equation*}
    p(g|\mu, \kappa) = \frac{\exp\left(\kappa \cos(x-\mu)\right)}{2 \pi I_0(\kappa)},
\end{equation*}

\noindent
where $I_0(\cdot)$ is the Bessel function and $\kappa$ plays a role similar to the inverse of the covariance in a Gaussian distribution. We can also define a mixture of two von Mises distributions, \revisedtext{that is}, $p(g) = \frac{1}{2}p(g|\mu_1, \kappa_1) + \frac{1}{2}p(g|\mu_2, \kappa_2)$. Our analysis in Fig.~\ref{fig:von_mise} shows that harmonic exponential distributions have a smaller $D_{KL}$ across a range of situations. When both approaches use few parameters, the harmonic exponential distribution does a better job of representing the underlying \revisedtext{density}. Increasing the number of parameters improves the fidelity of both approaches; however, the harmonic exponential distribution maintains its advantage in terms of $D_{KL}$. 

The parameter \(\kappa\) illustrates the \revised{``low-pass filter''} effect of the FFT: as \(\kappa\) increases, the distribution gains high-frequency components, leading to a rise in \(D_{KL}\), though, as Fig.~\ref{fig:von_mise} shows, the histogram approach performs worse. If the underlying density were \revised{discontinuous}, like a sum of Haar wavelets, the histogram approach might outperform the HEF in terms of $D_{KL}$. However, since many robotics problems involve smoothly varying functions, the HEF is better suited to handle continuous densities, as shown in this paper.

\begin{figure}[t]
\vspace{0.65em}
\centering
\begin{subfigure}[b]{0.326\columnwidth}
\includegraphics[width=\textwidth]{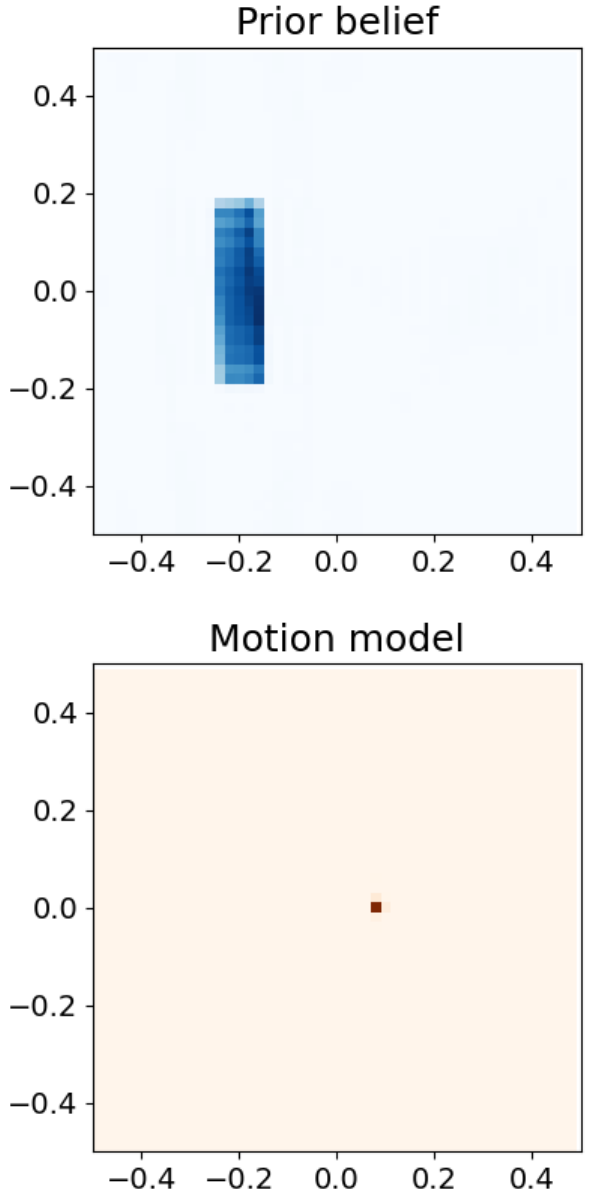}
\subcaption{}
 \label{fig:banana_a}
\end{subfigure} 
\begin{subfigure}[b]{0.65\columnwidth}
\includegraphics[width=\textwidth]{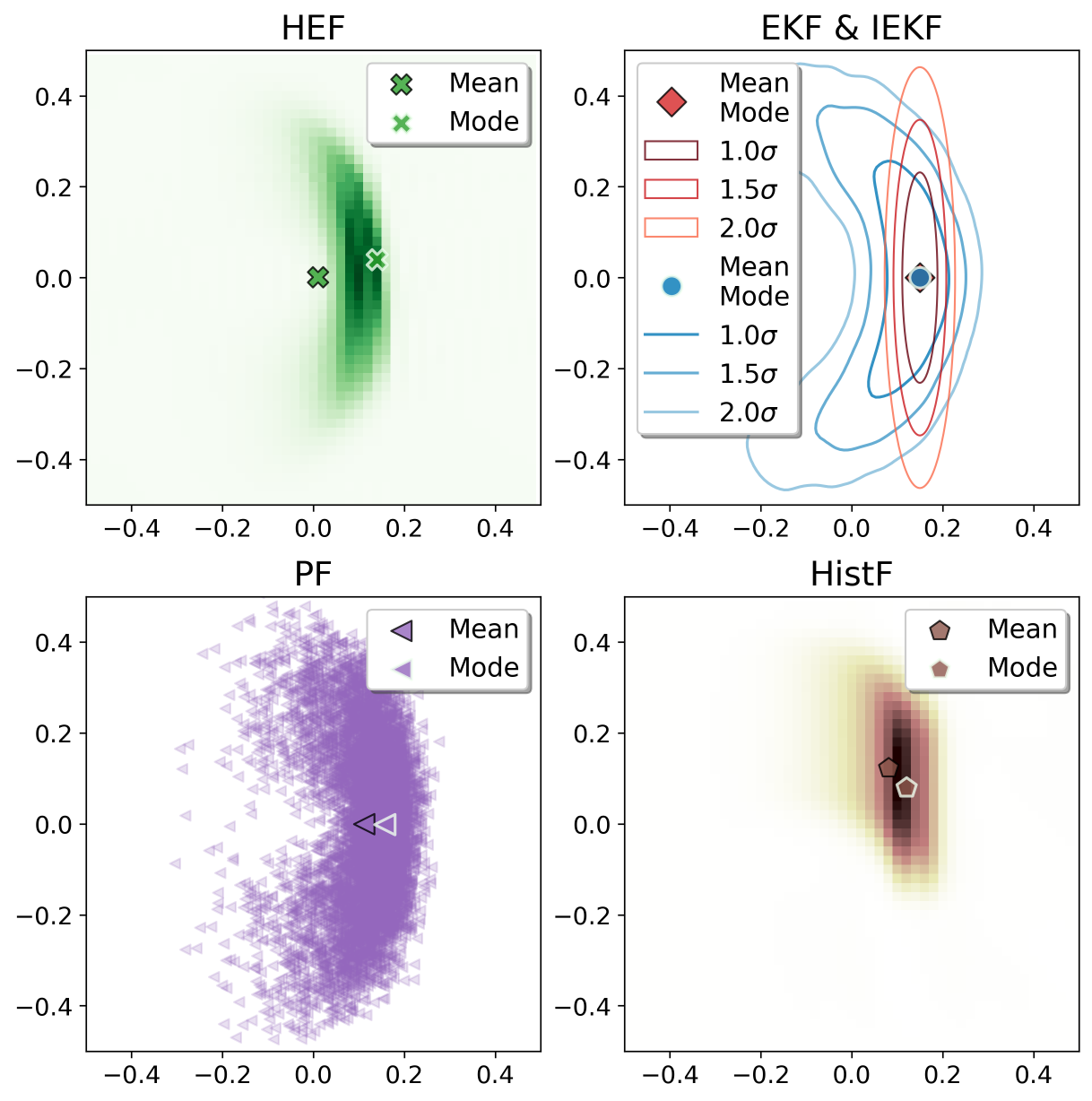}
\caption{}
\end{subfigure}
\caption{Ability of the HEF and other filtering approaches to represent the banana distribution. (a) shows the prior and \revisedtext{motion model}, which are modeled with a rectangular and Gaussian distribution, respectively. Both EKF \reviewed{and IEKF} use a Gaussian prior. (b) presents the posterior distribution of the four approaches after five prediction steps. \reviewed{The HEF, PF, and IEKF} are able to capture the banana distribution. }
\label{fig:banana_2}
\vspace{-1.0em}
\end{figure}

Another benefit of the HEF is its expressivity in capturing \revisedtext{diverse} distributions, such as the banana shape distribution presented by Thurn \emph{et al.} \revisedtext{\cite[Chapter 5]{thrun2005probrob}}. This distribution appears when the uncertainty in the heading of an agent increases due to the accumulation of error during multiple prediction steps. The standard EKF fails to capture this distribution, as it models uncertainty with error ellipses, which is inconsistent with the density of the banana distribution. Long \emph{et al.} \cite{long2012banana} showed that a Gaussian density on the Lie algebra can effectively represent the banana distribution, as shown in Fig.~\ref{fig:banana_2}. However, a Gaussian density in the Lie algebra can only model unimodal distributions.

We compare the expressivity of our method against \reviewed{four} other well-established filtering approaches by taking five prediction steps in a horizontal straight path to the right. Interestingly, \reviewed{the HEF, particle filter (PF) and IEKF} are able to capture the banana distribution. Note that all filters start from a rectangular prior, as shown in Fig.~\ref{fig:banana_a}, except for the EKF and IEKF due to its Gaussian assumptions. Although histogram filters (HistF) are highly expressive, they fail to capture the banana distribution, as 3D convolutions fail to model the effect of rotations in $SE(2)$.

\section{Evaluation} \label{sec:eval}

To evaluate our method, we use one simulated and two real-world datasets where the robot state is represented in $SE(2)$. Our assessment focuses on two main metrics. \reviewedtext{The first metric is the absolute trajectory error (ATE), calculated using two different estimators of the robot position: the mode and mean of the posterior belief. The ATE, measured in meters, is computed along the entire trajectory.} However, as this metric does not fully capture the non-Gaussian nature of multimodal estimators, we also calculate the negative \reviewed{log-posterior (NLP)} of the ground truth position under the posterior belief. We use the equation \reviewedtext{$\text{NLP} = -\frac{1}{T}\sum_t \log bel(x_t^{GT})$} where $T$ is the total number of time steps and \reviewedtext{$bel(x_t^{GT})$} is the posterior belief of the true state position, \reviewedtext{$x_t^{GT}$}, computed at time $t$. \revisedtext{The result of the HEF is the posterior belief over the state space which can be used in a state-belief planner \cite{platt2010belief}. \reviewed{NLP} helps to determine how useful our approach would be when used with these systems.}

To compute the FFTs, we define a \(50 \times 50 \times 32\) grid in \(SE(2)\), with \(50 \times 50\) Cartesian samples over \((x, y) \in [-0.5, 0.5]^2\) and 32 angular samples over \(\theta \in [0, 2\pi)\). The experiment maps are scaled to fit this grid. \reviewed{All HEF experiments use \( \lambda_{\text{max}} = 35 \), \revised{where the inverse transform \eqref{eq:invfft} is approximated using 111 samples}, \( m = 669 \), and \( n = 32 \).} We approximate the mean belief relying on the same samples used to calculate the FFTs, that is, \reviewed{$\mu \approx \sum_{n}bel(x_n)x_n$}. \reviewedtext{The mode of the HEF is also approximated using the FFT samples: $x^{*} = \underset{x \in {x_1, x_2, \dots, x_N}}{\argmax} \mathcal{F}^{-1} [\eta_{bel}](x)$.} 

In these assessments, our proposed HEF presented in Sec.~\ref{sec:method} is compared against the following filtering approaches: an EKF implementation that assumes unimodal Gaussian state and error, a histogram filter (HistF), and a particle filter (PF). For the sake of comparison, HEF and HistF use $80,000$ parameters, \reviewedtext{that is}, a $50\times50\times32$ grid \reviewedtext{and PF uses} $80,000$ particles. \reviewedtext{To determine the \reviewed{NLP} of the PF we approximate the density function using a histogram whose resolution is the same one used for the HEF and HistF.} \reviewedtext{The mode of the PF is estimated by taking the heaviest weighted particle before resampling}. For all our experiments we use as a motion model the one of a differential drive robot corrupted with Gaussian noise. To estimate the motion and measurement noise of each filter, we sweep over different noise values and took those that minimize the \reviewed{NLP}.

\subsection{Simulation}

\begin{figure}[t]
\vspace{0.65em}
\centering
\begin{subfigure}[b]{0.71\columnwidth}
\includegraphics[width=\textwidth]{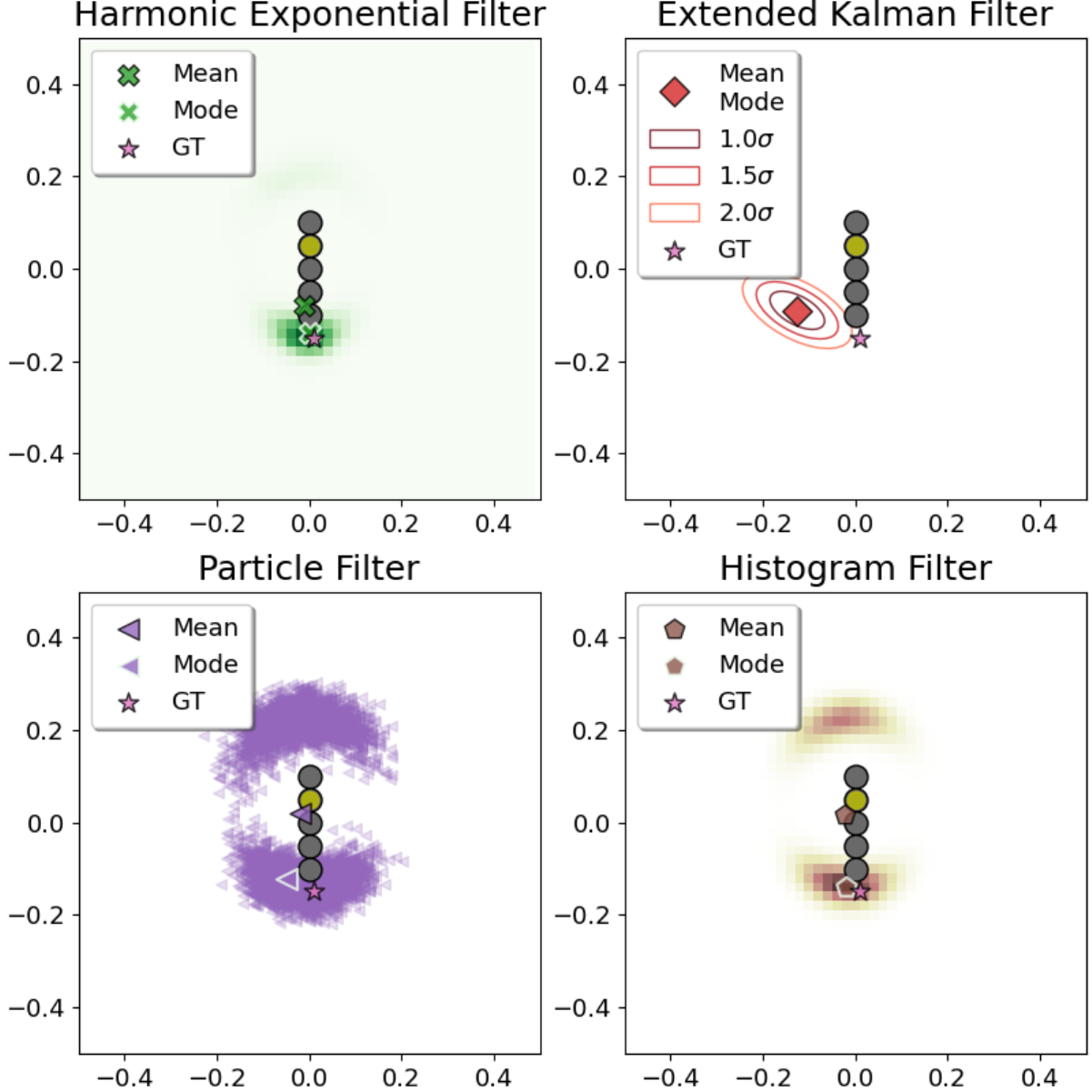}
\subcaption{}
\end{subfigure}
\hspace{0.5em}
\begin{subfigure}[b]{0.237\columnwidth}
\includegraphics[width=\textwidth]{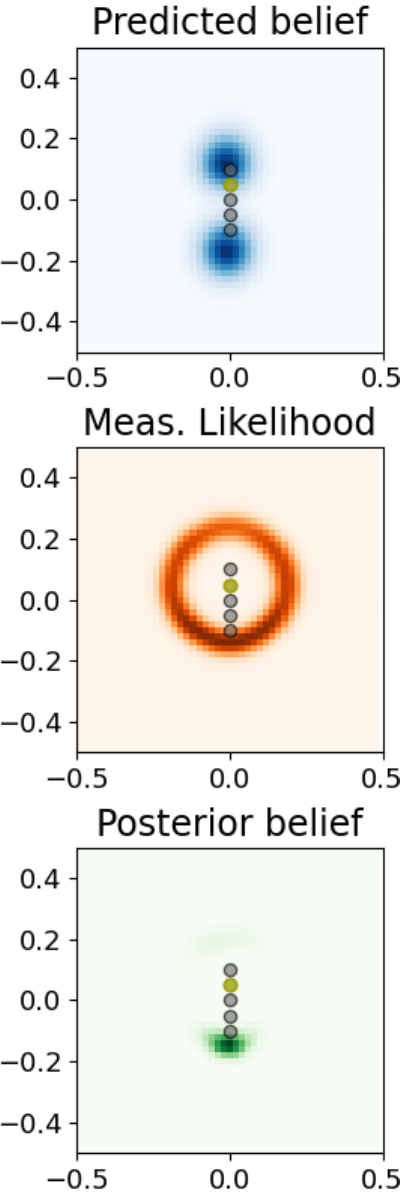}
\caption{}
\end{subfigure}
\caption{Sample results from the simulated environment. (a) shows the posterior belief for the baselines and the HEF. (b) depicts the steps of the HEF: the \revisedtext{predicted belief} (top) is $\overline{bel}_t$ from \eqref{eq:predict}, the range-only measurement likelihood (middle) is $p_{z_t \vert x_t}$, and the posterior belief (bottom) is $bel_t$ from \eqref{eq:udpate}.}
\label{fig:sim}
\end{figure}

The first evaluation is performed in a simulation environment with range-only measurements. This environment includes landmarks arranged in a line. The robot travels in a counterclockwise circle around the landmarks, which alternate in providing a range measurement at each time step, while assuming known correspondence. We chose to simulate range measurements due to their highly non-Gaussian measurement likelihood, which produces a ring around each landmark. \reviewed{All filters start from an equally-weighted bi-modal prior, as shown in Fig. \ref{fig:sim}. The EKF uses the weighted mean and covariance of this mixture as its prior.} We present the results of this experiment in Table \ref{table:sim_results}, \revisedtext{where metrics are computed over 10 different random seeds}. Overall, our method outperforms the baselines, obtaining the lowest ATE and \reviewed{NLP} in this multimodal localization experiment.

\begin{table}[ht]
\centering
\setlength{\tabcolsep}{5.0pt}
\begin{small}
\begin{tabular}{=l+r+r+r}
\toprule
& \multirowcell{2}{ATE w/ \\ mode (m)} & \multirowcell{2}{ATE w/ \\ mean (m)} &  \multirowcell{2}{NLP \\  (-)} \\ \\
\midrule
EKF & 0.101 $\pm$ 0.03 & 0.101 $\pm$ 0.03  & 271.2 $\pm$ 23.06 \\
HistF &  0.151 $\pm$ \textbf{0.02} & 0.132 $\pm$ \textbf{0.02}  & 17.04 $\boldsymbol{\pm}$ \textbf{0.57} \\
PF &  0.131 $\pm$ 0.03  & 0.102 $\pm$ \textbf{0.02} & 5.521 $\pm$ 1.15 \\
HEF (ours) &  \textbf{0.089} $\pm$ 0.03 & \textbf{0.082} $\boldsymbol{\pm}$ \textbf{0.02} & \textbf{2.242} $\pm$ 1.11 \\
\bottomrule
\end{tabular}
\end{small}
\caption{Quantitative results for the range-only simulator. Reported metrics include the absolute trajectory error (ATE) computed using the mode and the mean of the posterior belief, and the negative log-posterior (\reviewed{NLP}). Lower values are better. We report the mean and standard deviation of these metrics over 10 different random seeds.}
\label{table:sim_results}
\end{table}

\subsection{UWB Range Dataset}
\label{subsec:uwb}

\begin{figure}[ht]
\vspace{0.65em}
\centering
\begin{subfigure}[b]{0.45\columnwidth}
\includegraphics[width=\textwidth]{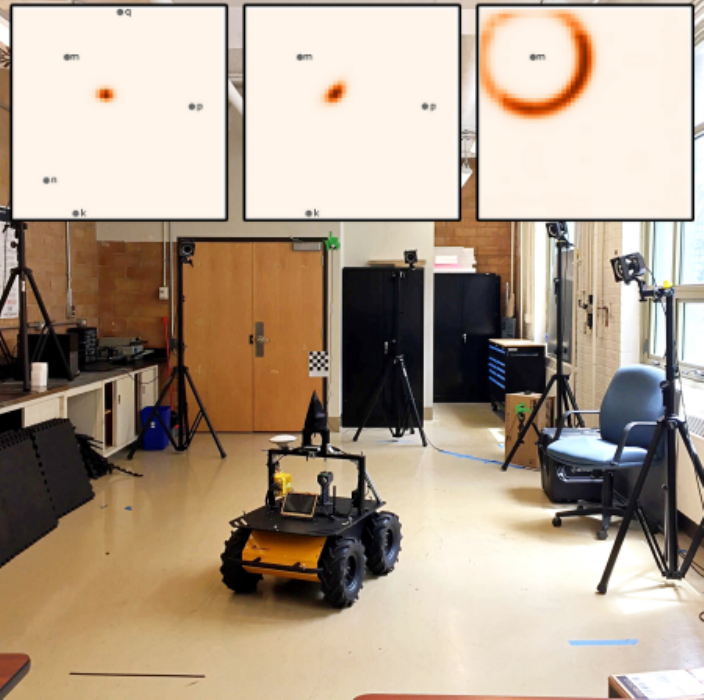}
\caption{UWB.}
\label{subfig:uwb}
\end{subfigure}
\hspace{1em}
\begin{subfigure}[b]{0.45\columnwidth}
\includegraphics[width=\textwidth,trim={0 0 0 35pt},clip]{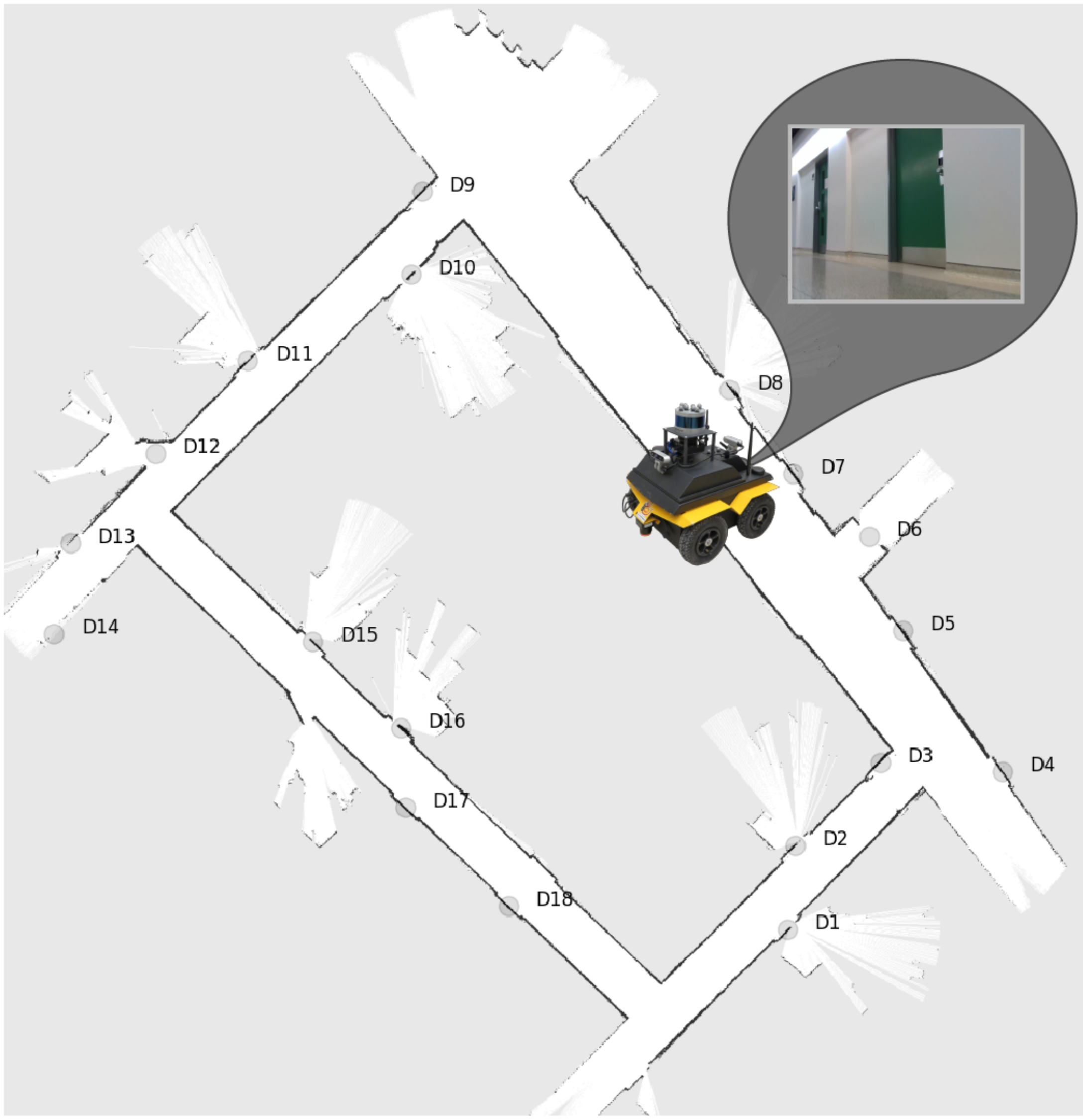}
\caption{Doors.}
\label{subfig:doors}
\end{subfigure}
\caption{Setup for the UWB and doors datasets. (a) shows the layout for the two UWB runs with a Clearpath Husky and five ultra-wideband (UWB) beacons. \reviewed{This image also provides a top-down view of the beacon configuration, showing how the measurement likelihood shifts from Gaussian to highly non-Gaussian as beacons are removed.}  (b) presents a top-down view of the laboratory used in the doors dataset.}
\label{fig:setup}
\vspace{-1.0em}
\end{figure}

For our first real-world evaluation, we used the Clearpath Husky unmanned ground vehicle (UGV) shown in Fig.~\ref{subfig:uwb}. The UGV is equipped with an ultra-wideband (UWB) tag that communicates with stationary UWB beacons in the surrounding room. Using a ranging protocol between the UGV-mounted UWB tag and the stationary UWB beacons, range measurements between each are computed. The experiment consists of two runs, each with five beacons that were mapped prior to the experiment. For each run, we removed beacons one by one ignoring their range measurements, producing five trials per run. We assume known correspondence. 

The motion of the agent was determined by integrating the IMU and wheel encoders of the Husky platform. \reviewedtext{The motion noise was estimated to have a standard deviation of 3.08 cm and 0.57 degrees per step}. Finally, the ground truth was provided by a NaturalPoint OptiTrack motion capture system. We repeat each run with a different number of beacons to demonstrate the robustness of our proposed HEF to handle unimodal and multimodal distributions. The layout of the beacons used in the experiments is shown in Fig.~\ref{subfig:uwb}. 

\begin{table}[!htpb]
\centering
\setlength{\tabcolsep}{2.75pt}
\begin{small}
\begin{tabular}{=l+r+r+r+r+r+r}
\toprule
& \multicolumn{3}{c}{5 Beacons} & \multicolumn{3}{c}{1 Beacon}  \\
\cmidrule(lr){2-4} \cmidrule(lr){5-7} 
& \multirowcell{3}{ATE w/ \\ mode \\ (m)}  & \multirowcell{3}{ATE w/ \\ mean \\ (m)} & \multirowcell{3}{\reviewed{NLP} \\ (-)}  & \multirowcell{3}{ATE w/ \\ mode \\ (m)}  & \multirowcell{3}{ATE w/ \\ mean \\ (m)} & \multirowcell{3}{\reviewed{NLP} \\ (-)}  \\ \\ \\
\midrule
EKF & \textrm{60161} & \textrm{60161} &\textrm{861e8} & \textrm{\textbf{0.903}} & \textrm{\textbf{0.903}} & \textrm{11.030} \\
HistF &  \textrm{0.221} & \textrm{0.210} & \textrm{-2.653} & \textrm{1.157} & \textrm{1.202} & \textrm{-0.364} \\
PF &  \textrm{0.304} & \textrm{0.210} & \textrm{3.311} & \textrm{3.731} & \textrm{2.639} & \textrm{9.253} \\
HEF (ours) &  \textrm{\textbf{0.199}} & \textrm{\textbf{0.194}} & \textrm{\textbf{-2.984}} & \textrm{1.616} & \textrm{1.076} & \textrm{\textbf{-0.533}} \\
\bottomrule
\end{tabular}
\end{small}
\caption{Quantitative results for the UWB dataset, \reviewedtext{run \RomanNumeralCaps{1}}. \reviewedtext{Reported metrics include the absolute trajectory error (ATE) computed using the mode and the mean of the posterior belief, and the negative log-posterior (\reviewed{NLP}). Lower values are better.} Note that the EKF is highly uncertain about the ground truth position of the agent during the experiments. \protect \footnotemark
}
\label{table:uwb}
\end{table}

The results of this evaluation are presented in Table~\ref{table:uwb}. In all runs and with different beacon configurations, the HEF is comparable to or better than the baseline filters even in the presence of a highly non-Gaussian measurement likelihood. Although the EKF obtained the best \reviewedtext{ATE} among all filters in the first run with one beacon, its \reviewed{NLP} is the largest among all filters, indicating that throughout the experiment the filter is highly uncertain about the ground truth position. \revisedtext{As all the results of the second run closely resembled those of the first run, they were not included in the table.}

\footnotetext{Complete results at \url{https://montrealrobotics.ca/hef/}.}

\subsection{Doors Dataset}

In our final real-world experiment, we draw inspiration from the semantic localization problem discussed in Thurn~\emph{et al.} \revisedtext{\cite[Chapter 1]{thrun2005probrob}}. The experiment involves localizing an agent navigating anticlockwise in a corridor using doors detected from RGB images as landmarks. All doors share similar semantic features, such as color or handle. Therefore, we do not assume known correspondence for which of the 21 doors generated a given measurement in this problem. 

We premapped the environment using gmapping \cite{Gmapping} and assumed a bearing-only measurement model that enforces a multimodal distribution. Bearing measurements are computed using the centroid of the doors. Bearing measurements are taken only for samples in the free space of the occupancy grid map. As in Sec.~\ref{subsec:uwb}, we integrate the IMU and wheel encoders to estimate the motion of a Jackal platform. \reviewedtext{The motion noise was estimated to have a standard deviation of 3.51 cm and 2.50 degrees per step}. The ground truth position was estimated using the laser-based SLAM implementation of gmapping. \revisedtext{The estimated entropy of the posterior distribution over the ground truth's pose using gmapping was $2.232$ nats.} The layout of the experiment, as well as an image of a door within the dataset is shown in Fig.~\ref{subfig:doors}.

\begin{table}[t]
    \medskip
    \centering
    \setlength{\tabcolsep}{5.0pt}
    \begin{small}
    \begin{tabular}{=l+r+r+r+r}
    \toprule
    & \multirowcell{2}{ATE w/ \\ mode (m)} & \multirowcell{2}{ATE w/ \\ mean (m)} &  \multirowcell{2}{NLP \\  (-)} \\ \\
    \midrule
    EKF & 2.224 $\pm$ 1.91 & 2.224 $\pm$ 1.91  & 521.479 \\
    HistF &  9.739 $\pm$ 7.78 & 8.806 $\pm$ 7.12  & 16.719  \\
    PF &  5.660 $\pm$ 4.78  & 3.541 $\pm$ 2.82 & 4.605 \\
    HEF (ours) &  \textbf{1.525} $\pm$ \textbf{1.34} & \textbf{1.559} $\boldsymbol{\pm}$ \textbf{1.36} & \textbf{0.986} \\
    \bottomrule
    \end{tabular}
    \end{small}
    \caption{Quantitative results for the doors dataset. \reviewedtext{Reported metrics include the absolute trajectory error (ATE) computed using the mode and the mean of the posterior belief, and the negative log-posterior (\reviewed{NLP}). \reviewedtext{We also report the standard deviation of the translation error.} }
    }
    \label{table:doors}
\end{table}

\begin{figure}[t]
\vspace{0.65em}
\centering
\begin{subfigure}[b]{0.71\columnwidth}
\includegraphics[width=\textwidth]{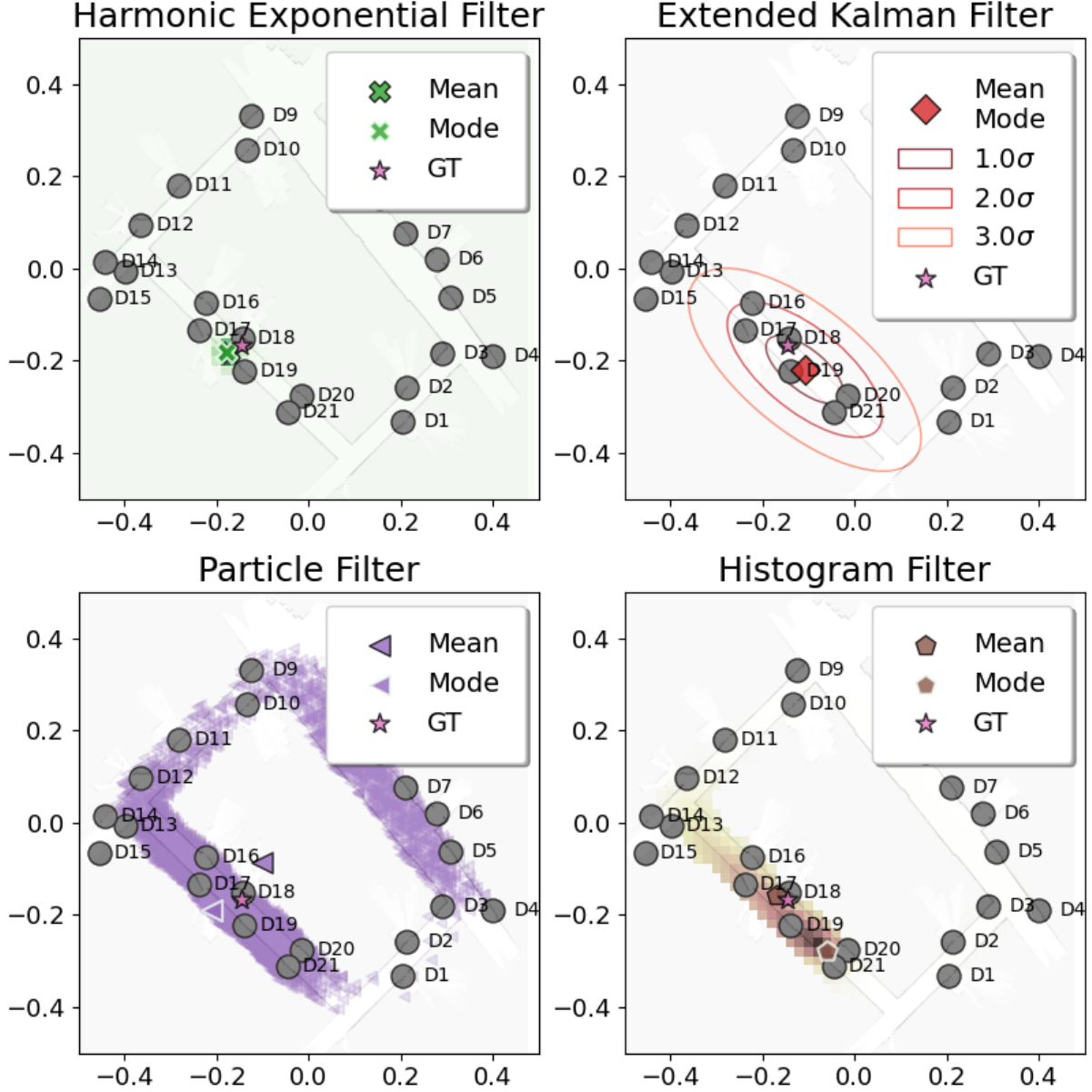}
\subcaption{}
\end{subfigure}
\hspace{0.5em}
\begin{subfigure}[b]{0.237\columnwidth}
\includegraphics[width=\textwidth]{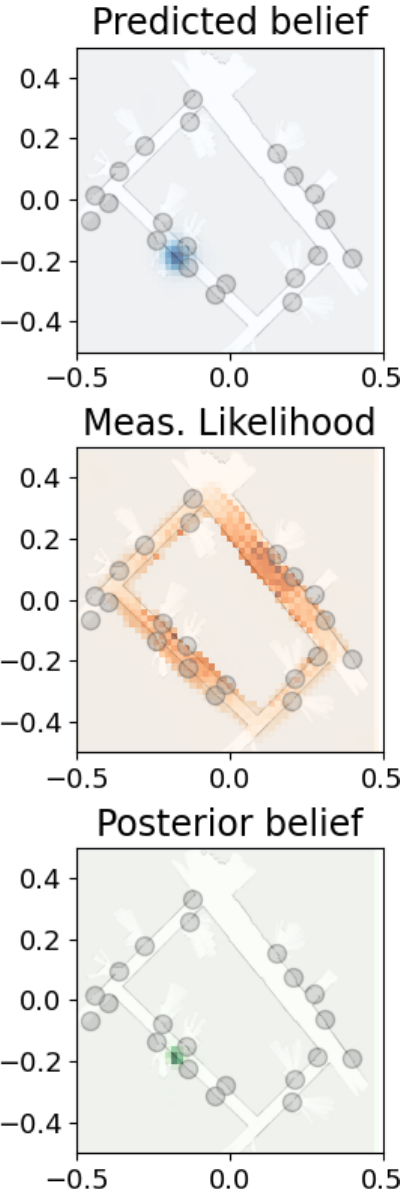}
\caption{}
\end{subfigure}
\caption{Sample results from the doors dataset. (a) shows the posterior belief for the baselines and the HEF. (b) features the steps of the HEF: the \revisedtext{predicted belief} (top) is $\overline{bel}_t$ from \eqref{eq:predict}, the bearing-only measurement likelihood (middle) is $p_{z_t \vert x_t}$, and the posterior belief (bottom) is $bel_t$ from \eqref{eq:udpate}.}
\label{fig:doors}
\end{figure}

As there is no known correspondence, we employ a greedy data association strategy, computing the measurement likelihood using only the most likely door for each measurement. We present the results of this experiment in Table~\ref{table:doors}. Our HEF surpasses by a large margin all baselines in this challenging task, showcasing its ability to handle multimodal distributions and the importance of symmetries for robot localization. A snapshot of the filters is shown in Fig.~\ref{fig:doors}.

During these experiments, our proposed approach took an average of 1.14s to run for each timestep. Our approach was implemented in python and ran on an i7-12700H CPU with 16GB of RAM. 

\section{Conclusion}\label{sec:discussion}

In this paper, we have presented a novel approach to nonparametric Bayesian filtering using the harmonic exponential distribution. We showed how our method was capable of capturing multimodal distributions while preserving the structure of non-commutative motion groups by relying on harmonic analysis \cite{chirikjian2001engineering}. Harmonic analysis on groups and homogeneous spaces has attracted new interest due to the growing field of geometric deep learning. As work in this field shows, FFT operations and related convolution computations are fully differentiable \cite{cohen2018spherical}. This leads to the possibility of using the harmonic exponential filter as a fully differentiable filter for end-to-end training of measurement and motion models. Another promising avenue for exploration is to use harmonic exponential distributions in a SLAM framework \revisedtext{or a state-belief planner \cite{platt2010belief}}. As Fourie \emph{et al.}~\cite{fourie2016bayestree} showed, all that is needed to perform belief propagation in a Bayes tree is the product operation between two distributions.

\bibliographystyle{IEEEtran}
\bibliography{IEEEabrv.bib,  short-string.bib, references.bib}

\end{document}